\documentclass[runningheads]{llncs}
\usepackage{graphicx}
\usepackage{epsfig}
\usepackage{comment}
\usepackage{amsmath,amssymb} 
\usepackage{color}
\usepackage{array}
\usepackage[noend]{algpseudocode}
\usepackage{algorithmicx,algorithm}
\usepackage{multirow}
\usepackage{subfigure}
\usepackage[pagebackref=true,breaklinks=true,letterpaper=true,colorlinks,bookmarks=false]{hyperref}
\usepackage{multirow}

\usepackage{tikz}
\usepackage{comment} 
\usepackage{amsmath,amssymb} 
\usepackage{color}

\begin{document}
\pagestyle{headings}
\mainmatter
\def\ECCVSubNumber{4118}  

\title{Improving the Transferability of Adversarial Examples with Resized-Diverse-Inputs, Diversity-Ensemble and Region Fitting} 

\titlerunning{Improving the Transferability with RDIM, DEM and RF}
%
\author{Junhua Zou\inst{1}\orcidID{0000-0003-4655-7173} \and
Zhisong Pan\inst{1}\orcidID{0000-0001-8615-7313} \and
Junyang Qiu\inst{2}\orcidID{0000--0001-8288-6180}\and
Xin Liu\inst{1}\orcidID{0000-0003-3051-4793}\and
Ting Rui\inst{1}\orcidID{0000--0002-2949-5874}\and
Wei Li\inst{1}\orcidID{0000-0003-1733-7956}}
\authorrunning{J. Zou et al.}
%
\institute{Army Engineering University of PLA, China \and
Jiangnan Institute of Computing Technology, China\\
\email{278287847@qq.com}}
\maketitle

\begin{abstract}
We introduce a three stage pipeline: resized-diverse-inputs (RDIM), diversity-ensemble (DEM) and region fitting, that work together to generate transferable adversarial examples. We first explore the internal relationship between existing attacks, and propose RDIM that is capable of exploiting this relationship. Then we propose DEM, the multi-scale version of RDIM, to generate multi-scale gradients. After the first two steps we transform value fitting into region fitting across iterations. RDIM and region fitting do not require extra running time and these three steps can be well integrated into other attacks. Our best attack fools six black-box defenses with a 93\% success rate on average, which is higher than the state-of-the-art gradient-based attacks. Besides, we rethink existing attacks rather than simply stacking new methods on the old ones to get better performance. It is expected that our findings will serve as the beginning of exploring the internal relationship between attack methods. Codes are available at https://github.com/278287847/DEM.


\end{abstract}

\section{Introduction}

\begin{figure}[!t]
 \centering
\includegraphics[width=0.9\textwidth]{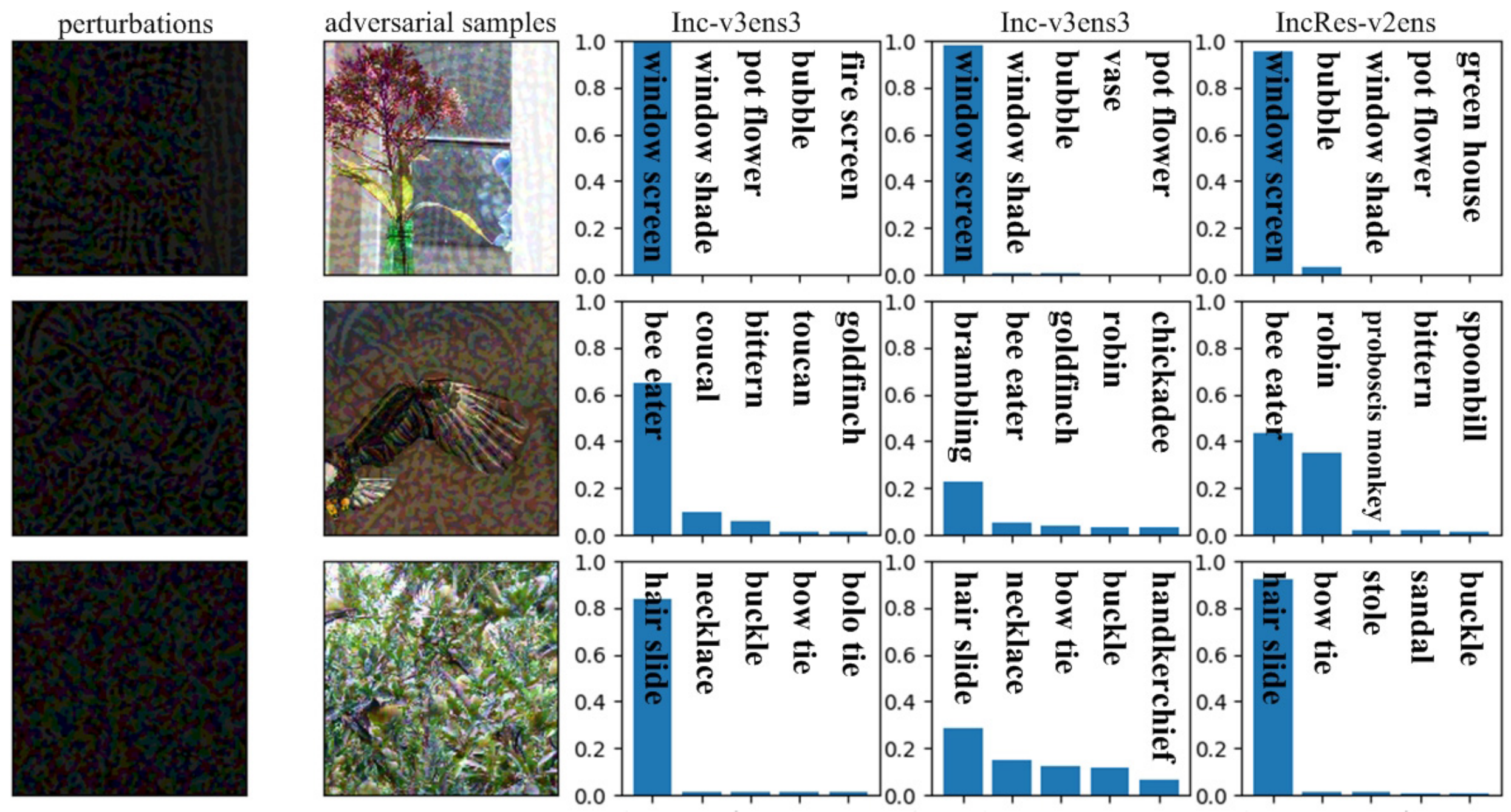}
\caption{Perturbations and adversarial samples along their predicted confidence scores of 3 black-box models. Inc-v3ens3, Inc-v3ens4 and IncRes-v2ens \cite{DBLP:conf/iclr/TramerKPGBM18} are defense models}
\label{fig:1}
\end{figure}

Recent work has demonstrated that deep neural networks (DNNs) are challenged by their vulnerability to adversarial examples \cite{DBLP:journals/corr/GoodfellowSS14,DBLP:journals/corr/SzegedyZSBEGF13}, i.e., inputs with carefully-crafted perturbations that are almost indistinguishable from the original images can be misclassified by DNNs. Moreover, a more severe and complicated security issue is the transferability of adversarial examples \cite{DBLP:conf/iclr/LiuCLS17,DBLP:conf/cvpr/Moosavi-Dezfooli17}, i.e., adversarial examples generated by a given DNN can also mislead other unknown DNNs. Fig.~\ref{fig:1} shows the transferability of an adversarial example. The threat of adversarial examples can even extend to the physical world \cite{DBLP:conf/icml/AthalyeEIK18,DBLP:conf/cvpr/EykholtEF0RXPKS18,DBLP:conf/iclr/KurakinGB17a}, and has motivated extensive research on security-sensitive applications. These defenses include adversarial training \cite{DBLP:journals/corr/GoodfellowSS14,DBLP:conf/iclr/MadryMSTV18,DBLP:conf/iclr/TramerKPGBM18}, input denoising \cite{DBLP:conf/cvpr/LiaoLDPH018}, input transformation \cite{DBLP:conf/iclr/GuoRCM18,DBLP:conf/iclr/XieWZRY18}, theoretically-certified defenses \cite{DBLP:conf/iclr/RaghunathanSL18,DBLP:conf/icml/WongK18} and others \cite{DBLP:conf/icml/PangDZ18,DBLP:conf/iclr/SamangoueiKC18}. Although adversarial examples are security threats to the practical deployment, they can help DNNs to identify the vulnerability before they are applied in reality \cite{DBLP:conf/iclr/MadryMSTV18}.

We focus on the gradient-based attacks of the classification task in this paper. With the knowledge of a given DNN, the gradient-based attacks are the most commonly used methods, and can attack black-box models based on the transferability of adversarial examples. Usually, existing attack methods are combined together to achieve higher attack success rates.

\begin{figure*}[!t]
 \centering
\includegraphics[width=0.9\textwidth]{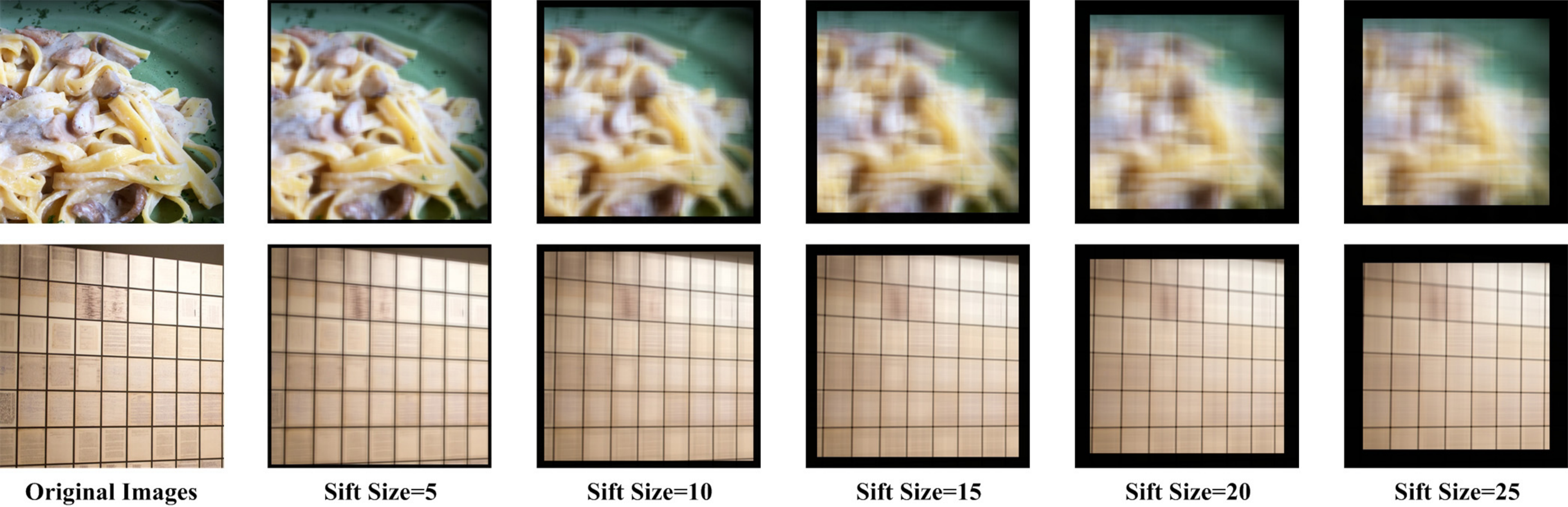}
\caption{Two rows of images generated by the translation-invariant method \cite{DBLP:conf/cvpr/DongPSZ19} with different sift size ranging from 5 to 25. Images of the first row gradually become blurred as the sift size increases while images of the second row remain stable}
\label{fig:2}
\end{figure*}

\begin{figure*}[!t]
 \centering
\includegraphics[width=0.9\textwidth]{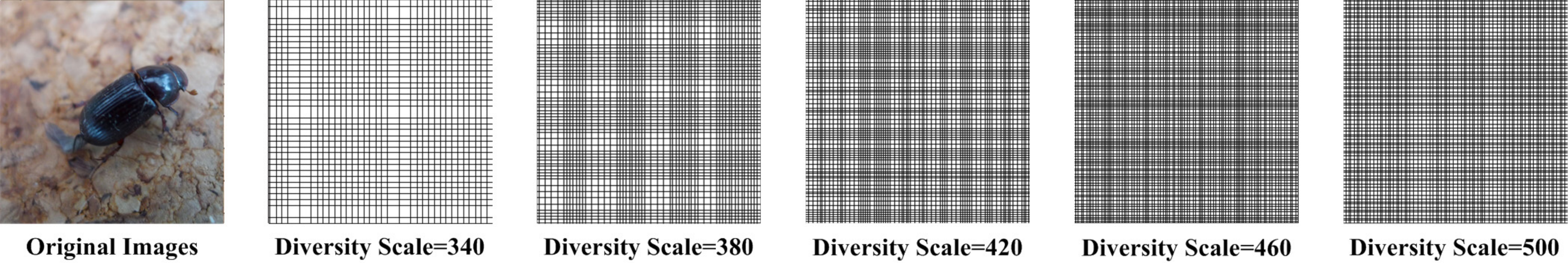}
\caption{A set of visualized gradients of a diverse example with different diversity scale \cite{DBLP:conf/cvpr/XieZZBWRY19}. The number of black stripes of these images increases as the diversity scale increases. In addition, gradients are visualized by setting non-zero values to 255 to highlight zero values. Hence, the black stripes indicate zero values of gradients while the white regions indicate non-zero values}
\label{fig:3}
\end{figure*}

\textbf{Motivation.} Among the gradient-based attacks, the diverse-inputs method (DIM) \cite{DBLP:conf/cvpr/XieZZBWRY19} applies random and differentiable transformations to the inputs with probability $p$, then feeds these transformed inputs into a white-box model for gradient calculation. Usually, DIM is combined with the translation-invariant method (TIM) \cite{DBLP:conf/cvpr/DongPSZ19} to achieve state-of-the-art results. Based on these two methods, our simple observations are shown as follow:
 \begin{itemize}
\item[1.] TIM can be considered as a Gaussian blur process for gradients. As shown in Fig.~\ref{fig:2}, TIM can blur a normal image (the first row), but cannot blur an image with vertical and horizontal stripes (the second row).
\item[2.] As shown in Fig.~\ref{fig:3}, the gradients of a diverse input have many vertical and horizontal stripes (here we visualize the gradients as images by setting non-zero values to 255 to highlight zero values). The number of stripes depends on the diversity scale.
\end{itemize}

Intuitively, DIM can alleviate the loss of gradient information caused by Gaussian blur, and thus generate more transferable adversarial examples. However, DIM sets up the transformation probability ${p}$ and limits the maximum diversity size to a really small size to avoid success rates dropping. The hyper-parameters of DIM restrict the number of stripes of the gradients, and cannot benefit TIM as much as possible. The intuition reveals the other two clues. One is that multi-scale gradient information benefits the transferability of adversarial examples. The other is that DIM divides the gradient information into many regions, and Gaussian Filter with large kernel may blur image edges. The two characteristics of these two methods indicate that region fitting plays a more important role than value fitting in adversarial example generation.

\textbf{Methods.} In this paper, we introduce a three stage pipeline: resized-diverse-inputs (RDIM), diversity-ensemble (DEM) and region fitting, that work together to generate transferable adversarial examples. We first explore the internal relationship between DIM \cite{DBLP:conf/cvpr/XieZZBWRY19} and TIM \cite{DBLP:conf/cvpr/DongPSZ19} based on the observations above, and propose a \textbf{resized-diverse-inputs method} (RDIM) that is more suitable to characterize this relationship. Compared with DIM, RDIM removes the transformation probability ${p}$, sets a much larger diversity size and finally resizes the diverse inputs to the original size at each iteration. We combine TIM and RDIM, and then conduct extensive experiments on the ImageNet dataset. The results show that this combination can achieve higher attack success rates on defense models comparing with the state-of-the-art results. We then propose a \textbf{diversity-ensemble method} (DEM), the multi-scale version of RDIM, to further boost the success rates. We show that DEM can further promote TIM because DEM generates multi-scale gradients with different numbers of vertical and horizontal stripes for TIM. After the first two steps we transform value fitting into \textbf{region fitting} across iterations. RDIM and region fitting do not require extra running time and these three steps can be well integrated into other attacks, such as model-ensemble methods \cite{DBLP:conf/cvpr/DongLPS0HL18}. Our best attack fools six black-box defenses with a 93\% attack success rate on average, which is higher than the state-of-the-art multi-model gradient-based attacks.

Rather than simply stacking the new methods on the old ones to get better performance, we rethink the proposed methods. It is expected that our findings will serve as the beginning of exploring the internal relationship between attack methods. In summary, our contributions are as follows:
\begin{itemize}
\item[1.] We are the first to explore the internal relationship between attack methods. We find that the gradients of diverse inputs have many vertical and horizontal stripes, and these gradients can be used to alleviate the loss of gradient information caused by TIM.
\item[2.] Based on the internal relationship between DIM and TIM, we propose RDIM to exploit this relationship. We show that RDIM further boosts the attack success rates against black-box defenses.
\item[3.] We propose DEM which can generate multi-scale gradients for TIM. DEM can further promote TIM because DEM generates multi-scale gradients with different numbers of vertical and horizontal stripes for TIM. We also transform value fitting into region fitting across iterations to further boost the success rates against black-box defenses.
\item[4.] Our best attack fools six black-box defenses with a 93\% attack success rate on average, which is higher than the state-of-the-art gradient-based attacks.

\end{itemize}

\section{Related Work}

Recent work has demonstrated that DNNs are challenged by their vulnerability to adversarial examples \cite{DBLP:conf/pkdd/BiggioCMNSLGR13,DBLP:journals/corr/SzegedyZSBEGF13}. The primary purposes of adversarial example generating methods are high attack success rates with minimal size of perturbations \cite{DBLP:conf/sp/Carlini017}. Attack methods in the classification task can be categorized into three types---the gradient-based attacks \cite{DBLP:conf/cvpr/DongLPS0HL18,DBLP:journals/corr/GoodfellowSS14,DBLP:conf/iclr/KurakinGB17}, the score-based attacks \cite{DBLP:journals/corr/NarodytskaK16} and the decision-based attacks \cite{DBLP:conf/iclr/BrendelRB18,DBLP:journals/corr/abs-1904-02144}. In addition, adversarial examples exist in face recognition \cite{DBLP:conf/mmsp/BoseA18}, object detection \cite{DBLP:conf/cvpr/EykholtEF0RXPKS18}, semantic segmentation \cite{DBLP:conf/cvpr/ArnabMT18}, etc.. In this paper, we focus on gradient-based attacks of the classification task.

Gradient-based attacks can be categorized into three parts---the gradient processing part, the ensemble part and the input preprocessing part. In the gradient processing part, Goodfellow et al. \cite{DBLP:journals/corr/GoodfellowSS14} proposed the fast gradient sign method (FGSM) to craft adversarial examples by performing one-step update efficiently. Kurakin et al. \cite{DBLP:conf/iclr/KurakinGB17a} extended FGSM to the basic iterative method (BIM) and showed the powerful ability of BIM in white-box attacks but lousy performance in black-box attacks. Dong et al. \cite{DBLP:conf/cvpr/DongLPS0HL18} proposed the momentum iterative fast gradient sign method (MI-FGSM) to boost success rates in black-box attacks by integrating a momentum term into BIM. Lin et al. \cite{DBLP:journals/corr/abs-1908-06281} proposed the Nesterov iterative fast gradient sign method (NI-FGSM) to further improve the transferability of adversarial examples by adapting Nesterov accelerated gradient into MI-FGSM. In the ensemble part, Dong et al. \cite{DBLP:conf/cvpr/DongLPS0HL18} proposed a model ensemble method to fool robust black-box models obtained by ensemble adversarial training. Lin et al. \cite{DBLP:journals/corr/abs-1908-06281} used a set of scaled images to achieve model augmentation and named it scale-invariant attack method (SIM). In the input preprocessing part, Dong et al. \cite{DBLP:conf/cvpr/DongPSZ19} proposed the translation-invariant attack method (TIM) to generate adversarial examples that are less sensitive to the discriminative regions. Xie et al. \cite{DBLP:conf/cvpr/XieZZBWRY19} proposed the input diversity (DIM) to generate adversarial examples by iteratively applying the random transformation to input examples.

Most of defenses can be categorized into two types---adversarial training and input modification. Adversarial training \cite{DBLP:journals/corr/GoodfellowSS14} mainly augmented the training dataset by its adversarial examples during the training process to broaden the discriminative regions \cite{DBLP:conf/iclr/KurakinGB17}. Additionally, Tram{\`{e}}r et al. \cite{DBLP:conf/iclr/TramerKPGBM18} further improved the robustness of defense models and proposed the ensemble adversarial training by augmenting clean examples with adversarial examples crafted for various models. Input modification aimed to reduce the influence of adversarial examples on models by mitigating adversarial perturbations through different modification methods. Xie et al. \cite{DBLP:conf/iclr/XieWZRY18} employed random resizing and padding to defense against the adversarial attacks. Liao et al. \cite{DBLP:conf/cvpr/LiaoLDPH018} reduced the effects of adversarial perturbations using high-level representation guided denoiser.

\section{Methodology}

Given an input example ${X}$ which we call a clean example, and it can be correctly classified to the ground-truth label ${{y}}_{true}$ by deep model $f(\cdot)$ to $f(X) = {y_{true}}$. It is possible to construct two types of adversarial examples to attack model $f(\cdot)$ by adding different adversarial perturbations to the clean example ${X}$. In non-targeted attack, an adversarial example ${X^{adv}}$ is generated with the ground-truth label ${{y}}_{true}$ to mistaken the model as $f({X^{adv}}) \ne {y_{true}}$. In targeted attack, a targeted adversarial example ${X^{adv}}$ is classified to the specified target class ${{y}}_{target}$ as $f({X^{adv}}) = {y_{target}}$, where ${y_{target}} \ne {y_{true}}$. In the standard case, in order to generate indistinguishable adversarial example ${X^{adv}}$, the distortion between adversarial example ${X^{adv}}$ and clean example ${X}$ is measured as ${L_p}$ norm of the adversarial noise as ${\left\| {{X^{adv}} - X} \right\|_p} \le \varepsilon$, where ${p}$ could be $0$, $1$, $2$, $\infty $, and $\varepsilon $ is the size of the adversarial perturbation.

\subsection{Gradient-Based Attack Methods}

In this subsection, we present a brief introduction of the family of the gradient-based attack methods.

\textbf{Fast Gradient Sign Method} (FGSM)  \cite{DBLP:journals/corr/GoodfellowSS14} generates an adversarial example ${X^{adv}}$ by maximizing the loss function $J\left( {{X^{adv}},{y_{true}}} \right)$ of a pre-trained DNN. FGSM can efficiently craft an adversarial example as
\begin{equation}\label{eq:4}
{X^{adv}} = X + \varepsilon  \cdot sign\left( {{\nabla _X}J\left( {X,{y_{true}}} \right)} \right),
\end{equation}
where ${\nabla _X}J\left( {\cdot,\cdot} \right)$ computes the gradient of the loss function w.r.t. $X$, $sign\left( \cdot \right)$ is the sign function, and $\varepsilon$ is the required scalar value that basically restricts the ${L_\infty}$ norm of the perturbation.

\textbf{Iterative Fast Gradient Sign Method} (I-FGSM)  \cite{DBLP:conf/iclr/KurakinGB17a} applies FGSM multiple times with a small steps size $\alpha$, while FGSM generates an adversarial example by taking a single large step in the direction. The basic iterative method (BIM) \cite{DBLP:conf/iclr/KurakinGB17a} starts with $X_0^{adv} = X$, and iteratively computes as
\begin{equation}\label{eq:6}
X_{t + 1}^{adv} =  {X_t^{adv} + \alpha  \cdot sign\left( {{\nabla _{X_t^{adv}}} J\left( {X_t^{adv},{y_{true}}} \right)} \right)},
\end{equation}
where ${X_t^{adv}}$ denotes the adversarial example generated at the ${t}$-th iteration, and ${X_0^{adv} = X}$. 

\textbf{Momentum Iterative Fast Gradient Sign Method} (MI-FGSM)  \cite{DBLP:conf/cvpr/DongLPS0HL18} enhances the transferability of adversarial examples in black-box attacks and maintains the success rates in white-box attacks. The updating procedures are
\begin{equation}\label{eq:9}
{g_{t + 1}} = \mu  \cdot {g_t} + \frac{{{\nabla _{X_t^{adv}}}J\left( {X_t^{^{adv}},{y_{true}}} \right)}}{{{{\left\| {{\nabla _{X_t^{adv}}}J\left( {X_t^{^{adv}},{y_{true}}} \right)} \right\|}_1}}},
\end{equation}
\begin{equation}\label{eq:10}
X_{t + 1}^{^{adv}} = {X_t^{^{adv}} + \alpha \cdot   sign\left( {{g_{t + 1}}} \right)},
\end{equation}
where ${g_t}$ denotes the accumulated gradient at the ${t}$-th iteration, and $\mu$ is the decay factor of ${g_t}$.

\textbf{Nesterov Iterative Fast Gradient Sign Method} (NI-FGSM)  \cite{DBLP:journals/corr/abs-1908-06281} integrates Nesterov accelerated gradient into gradient-based attack methods to avoid the “missing” of the global maximum as
\begin{equation}\label{eq:11}
X_t^{nes} = X_t^{adv} + \alpha \cdot \mu \cdot {g_t},
\end{equation}
\begin{equation}\label{eq:12}
{g_{t + 1}} = \mu \cdot {g_t} + \frac{{{\nabla _{X_t^{adv}}}J\left( {X_t^{nes},{y^{true}}} \right)}}{{{{\left\| {{\nabla _{X_t^{adv}}}J\left( {X_t^{nes},{y^{true}}} \right)} \right\|}_1}}},
\end{equation}
\begin{equation}\label{eq:13}
X_{t + 1}^{adv} = {X_t^{adv} + \alpha \cdot {\rm{sign}}\left( {{g_{t + 1}}} \right)}.
\end{equation}

\textbf{Diverse-Inputs Method} (DIM) \cite{DBLP:conf/cvpr/XieZZBWRY19} generates adversarial examples by applying the random transformation to input examples at each iteration where the transformation function ${T(X_t^{adv},p)}$ is
\begin{equation}\label{eq:14}
T(X_t^{adv},p)=\left\{\begin{array}{ll}
T(X_t^{adv}) & \text{with probability \textit{p}}\\
X_t^{adv} & \text{with probability 1-\textit{p}}\end{array} \right.
\end{equation}

\textbf{Translation-Invariant Method} (TIM)  \cite{DBLP:conf/cvpr/DongPSZ19} uses a set of translated images to form an adversarial example as
\begin{equation}\label{eq:15}
	\left.\begin{array}{ll}
	 X_{t+1}^{adv} = \sum_{i,j}T_{ij}(X_t^{adv}),\\
	  s.t.\,\,\|X_t^{adv} - X^{real}\|_{\infty} \leq \epsilon,
	\end{array}\right.
\end{equation}
where $T_{ij}(X_t^{adv})$ denotes the translation function that respectively shifts input $X_t^{adv}$ by $i$ and $j$ pixels along the two-dimensions.

\subsection{Observation Analyses}
\label{3.2}

Our simple observations are shown in Fig.~\ref{fig:2} and Fig.~\ref{fig:3}. TIM fails to blur an image with vertical and horizontal stripes, while the gradients of a diverse input have many vertical and horizontal stripes. Intuitively, DIM can alleviate the loss of gradient information caused by Gaussian blur, and thus generate more transferable adversarial examples. We present the analyses as follow:
\begin{itemize}
\item[1.] Compared with the normal size ($299 \times 299 \times 3$), the input of DIM is a larger example ($S_1 \times S_1 \times 3$, where $S_1>299$), which leads to the deviation of the model output. DIM does not resize the diverse inputs to the original size after the process. Hence, diversity scale of DIM is limited to 330 to avoid the vast size difference between the original inputs and the diverse inputs. Additionally, the probability $p$ of DIM also limits the diversity.
\item[2.] TIM can be considered as a Gaussian blur process and cause the loss of gradient information. Lin et al. \cite{DBLP:journals/corr/abs-1908-06281} show that TIM with a smaller kernel is better in multi-model attack. In this paper, we find another way to alleviate the loss of gradient information. Gaussian blur cannot blur an image with vertical and horizontal stripes while RDIM fills this gap.
\item[3.] Multi-group of gradients with different diversity scales can satisfy the need of TIM for blurring images with different types of stripes.
\item[4.] DIM divides the gradient information into many regions, and Gaussian Filter with large kernel may blur image edges. The two characteristics of these two methods indicate that region fitting plays a more important role than value fitting in adversarial example generation.

\end{itemize}
 
\subsection{Resized-Diverse-Inputs Method}

Based on the analyses above, we propose a \textbf{resized-diverse-inputs method} (RDIM) that is more suitable for the internal relationship with TIM \cite{DBLP:conf/cvpr/DongPSZ19}. Compared with DIM, RDIM removes the transformation probability ${p}$, sets a much larger diversity size and finally resizes the diverse inputs to the original size at each iteration. These three improvements of RDIM correspond to the first two analyses in Sec.~\ref{3.2}. The algorithm of the RDIM is presented in Algorithm~\ref{al:1}.

\begin{algorithm}[t]
\caption{RDIM}
\label{al:1}
{\bf Input:}
An example $X$; the original size $S$; the diversity scale $S_1$.\\
{\bf Output:}
A diverse example ${X^{d}}$.
\begin{algorithmic}[1]
\State $a \sim \text{Unif}\left( {S,{S_1}} \right)$; \/// get the random size $a$ 
\State ${X^r} = resize\left( {X,\left( {a,a} \right)} \right)$; \/// resize the input image to the random size $a$
\State $H = {S_1} - a$; \/// get the padding size $H$ 
\State $top, left \sim \text{Unif}\left( {0,H} \right)$; \/// get the random top and left padding size
\State $bottom = H - top, right = H-left$; \/// get the bottom and right padding size
\State ${X^p} = padding\left( {{X^r},\left( {top,bottom,left,right} \right)} \right)$; \/// get the padding image ${X^p}$
\State {\bf Return} ${X^{d}} = resize\left( {{X^p},\left( {S,S} \right)} \right)$. \/// resize the padding image to the original size
\end{algorithmic}
\end{algorithm}

\subsection{Diversity-Ensemble Method}

For multi-scale setting, we also propose a \textbf{diversity-ensemble method} (DEM), the multi-scale version of RDIM, to improve the transferability of adversarial examples. Inspired by the third analysis in Sec.~\ref{3.2}, we propose DEM, which generates multi-scale gradients with different numbers of vertical and horizontal stripes for TIM. DEM can satisfy the need of TIM for blurring images with different types of stripes. Similar to the ensemble-in-logits method \cite{DBLP:conf/cvpr/DongLPS0HL18}, we fuse the logits of $K$ diversity scales as
\begin{equation}\label{eq:16}
l\left( X \right) = \sum\nolimits_{k = 1}^K {{\omega _k}l\left( {T\left( {X,{S_k}} \right)} \right)},
\end{equation}
where ${l\left( {T\left( {X,{S_k}} \right)} \right)}$ denotes the logits of resized diverse inputs with $k_{th}$ scale, ${\omega _k}$ denotes the ensemble weight with ${\omega _k} \ge 0$ and $\sum\nolimits_{k = 1}^K {{\omega _k}} = 1$.

\subsection{Region Fitting}

\begin{figure*}[t]
 \centering
\includegraphics[width=0.95\textwidth]{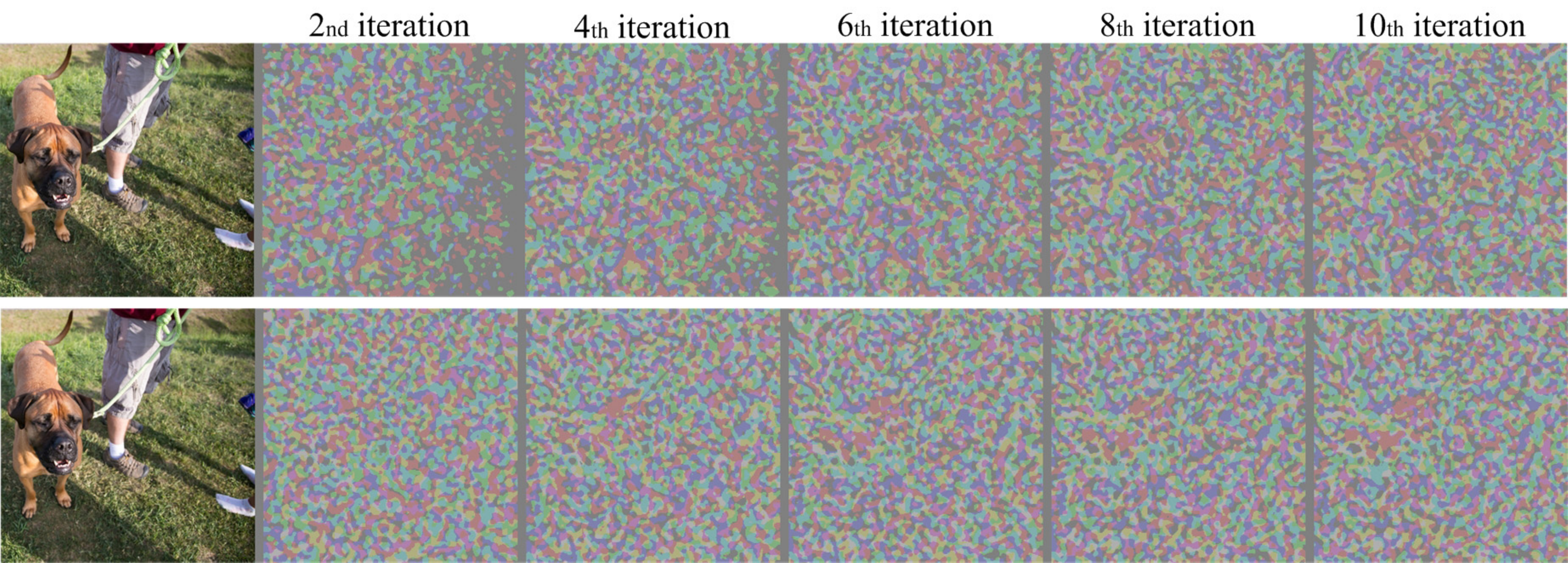}
\caption{Visualization of perturbations respectively generated with value fitting (the first row) and region fitting (the second row) in ten iterations. The value fitting cannot craft perturbations with detailed texture in the first four iterations}
\label{fig:5}
\end{figure*}

TIM can be considered as a Gaussian blur process with a large kernel ($15 \times 15$) for gradients while DIM divides the gradients into many regions. These two methods for gradients mainly process the pixel region while normal iterative methods fit pixel value iteratively. Hence, we transform value fitting into region fitting across iterations. Compared with the updating procedure Eq.~\eqref{eq:13}, region fitting can be expressed as

\begin{equation}\label{eq:31}
X_{t + 1}^{adv} = Clip_\varepsilon \left\{ {X_t^{adv} + \varepsilon  \cdot {\rm{sign}}\left( {{g_{t + 1}}} \right)} \right\}.
\end{equation}

The difference between Eq.~\eqref{eq:13} and Eq.~\eqref{eq:31} across iterations is that we change $\alpha$ into $\varepsilon$. Eq.~\eqref{eq:13} iteratively increases the perturbation size with step size $\alpha$, and finally makes the perturbation size reach $\varepsilon$. Eq.~\eqref{eq:31} makes the perturbation size reach $\varepsilon$ at each iteration, and generates adversarial examples to meet the ${L_\infty }$ norm bound by clipping function. Dong et al. \cite{DBLP:conf/cvpr/DongPSZ19} show that the classifiers rely on different discriminative regions for predictions. Region Fitting can accelerate the process of searching the discriminative regions as shown in Fig.~\ref{fig:5}.


We summarize RF-DE-TIM (the combination of TIM, RDIM, DEM, region fitting and MI-FGSM) in Algorithm~\ref{al:2}.

\begin{algorithm}[t]
\caption{RF-DE-TIM}
\label{al:2}
{\bf Input:}
A clean example $X$ and ground-truth label ${y_{true}}$; the logits of $K$ diversity scales $l\left( {T\left( {X,{S_1}} \right)} \right),l\left( {T\left( {X,{S_2}} \right)} \right), \ldots ,l\left( {T\left( {X,{S_K}} \right)} \right)$; ensemble weights ${\omega _1},{\omega _2}, \ldots ,{\omega _k}$;\\
{\bf Input:}
The perturbation size $\varepsilon$; iterations $T$ and decay factor $\mu$. \\
{\bf Output:}
An adversarial example ${X^{adv}}$.
\begin{algorithmic}[1]
\State $\alpha  = \varepsilon /T$;
\State ${g_0} = 0$; $X_{_0}^{adv} = X$;
\For{$t = 0$ to $T-1$}
\State Input $X_t^{adv}$;
\State Get logits $l\left( {X_t^{adv}} \right)$ by Eq.~\eqref{eq:16};
\/// fuse the logits of $K$ diversity scales
\State Get the gradient ${\nabla _X}J\left( {X_t^{adv},{y_{true}}} \right)$;
\State Process the gradient by $W * {\nabla _X}J\left( {X_t^{adv},{y_{true}}} \right)$; \/// Gaussian blur for gradient
\State Update ${g_{t + 1}}$ by Eq.~\eqref{eq:12}; \/// accumulate the gradient
\State Update $X_{t + 1}^{adv}$ by Eq.~\eqref{eq:31}; \/// apply the region fitting
\EndFor
\State {\bf Return} ${X^{adv}} = X_t^{adv}$.
\end{algorithmic}
\end{algorithm}

\section{Experiments}

To validate the effectiveness of our methods, we present extensive experiments on ImageNet dataset. Table~\ref{tab:10} introduces the abbreviations used in the paper.

We first provide experimental settings in Sec.~\ref{4.1}. Then we report the internal relationship between RDIM and TIM with the opposite results of different combinations of attack methods in Sec.~\ref{4.2}. Finally, we compare the results of our methods with the baseline methods in Sec.~\ref{s4.3} and Sec.~\ref{4.4}.

\subsection{Experimental Settings}
\label{4.1}

\textbf{Dataset.} We utilize an ImageNet-compatible dataset\footnote{\url{https://github.com/tensorflow/cleverhans/tree/master/examples/nips17_adversarial_competition/dataset}} \cite{DBLP:journals/ijcv/RussakovskyDSKS15} used in the NIPS 2017 adversarial competition to comprehensively compare the results of our methods with the baseline methods. The image size is $299 \times 299 \times 3$.

\begin{table*}[!t]
	\centering
	\caption{\label{tab:10}Abbreviations used in the paper}
	\begin{tabular}{c|c}
		\hline
		Abbreviation & Explanation\\
		\hline
		RDI-FGSM & The combination of RDIM and FGSM\\
		\hline
		RDI-MI-FGSM & The combination of RDIM and MI-FGSM\\
		\hline
		TI-RDIM & The combination of RDIM, TIM and MI-FGSM\\
		\hline
		TI-DIM & The combination of DIM, TIM and MI-FGSM\\
		\hline
		NI-TI-RDIM & The combination of RDIM, TIM and NI-FGSM\\
		\hline
		NI-TI-DIM & The combination of DIM, TIM and NI-FGSM\\
		\hline
		DE-TIM & The combination of RDIM, DEM, TIM and MI-FGSM\\
		\hline
		SI-TIM & The combination of SIM, DIM, TIM and MI-FGSM\\
		\hline
		DE-NI-TIM & The combination of RDIM, DEM, TIM and NI-FGSM\\
		\hline
		SI-NI-TIM & The combination of SIM, DIM, TIM and NI-FGSM\\
		\hline
		RF-TI-RDIM & The combination of region fitting, RDIM, TIM and MI-FGSM\\
		\hline
		RF-DE-TIM & The combination of region fitting, RDIM, DEM, TIM and MI-FGSM\\
		\hline
	\end{tabular}
\end{table*}

\textbf{Models.} We consider six defense models---Inc-v3ens3, Inc-v3ens4, IncRes-v2ens \cite{DBLP:conf/iclr/TramerKPGBM18}, high-level representation guided denoiser (HGD) \cite{DBLP:conf/cvpr/LiaoLDPH018}, input transformation through random resizing and padding (R\&P) \cite{DBLP:conf/iclr/XieWZRY18}, and rank-3 submission\footnote{\url{https://github.com/anlthms/nips-2017/tree/master/mmd}} in the NIPS 2017 adversarial competition, as the robust black-box defense models. To attack these models mentioned above, we also consider four normally trained models---Inception v3 (Inc-v3) \cite{DBLP:conf/cvpr/SzegedyVISW16}, Inception v4 (Inc-v4), Inception ResNet v2 (IncRes-v2) \cite{DBLP:conf/aaai/SzegedyIVA17} and ResNet v2-152 (Res-v2-152) \cite{DBLP:conf/eccv/HeZRS16}, as the white-box models to craft adversarial examples. It should be noted that adversarial examples crafted for four normally trained models are unaware of any defense strategies and will be used to attack six defense models, including top-3 defense solutions of NIPS 2017 adversarial competition.

\textbf{Baselines.} In our experiments, we first explore the internal relationship between attack methods by RDI-FGSM, RDI-MI-FGSM, and TI-RDIM . Then in single-scale attack manner, we respectively compare TI-RDIM and NI-TI-RDIM with two baseline methods, TI-DIM and NI-TI-DIM. In the multi-scale attack manner, we respectively compare DE-TIM and DE-NI-TIM with two baseline methods, SI-TIM and SI-NI-TIM. We also include RF-DE-TIM, SI-NI-TIM and SI-NI-TI-DIM \cite{DBLP:journals/corr/abs-1908-06281} in ensemble-based attacks for comparison.

\textbf{Hyper-parameters.} We follow the settings in TIM \cite{DBLP:conf/cvpr/DongPSZ19} with the number of iteration as $T = 10$, the maximum perturbation as $\varepsilon = 16$, the decay factor as $\mu = 1.0$. For TIM, We set the kernel size to $15 \times 15$. For SI-NI-FGSM, we follow the settings in NIM \cite{DBLP:journals/corr/abs-1908-06281} with the number of the scale copies as $m = 5$. For DEM, we set the diversity list to $[340,380,420,460,500]$. Please note that the hyper-parameters settings for all attacks are the same.

\subsection{The Internal Relationship}
\label{4.2}

\begin{figure*}[t]
\centering
\subfigure[RDI-FGSM]{
\begin{minipage}[b]{0.3\linewidth}
\label{fig:3.1}
\includegraphics[width=0.97\textwidth]{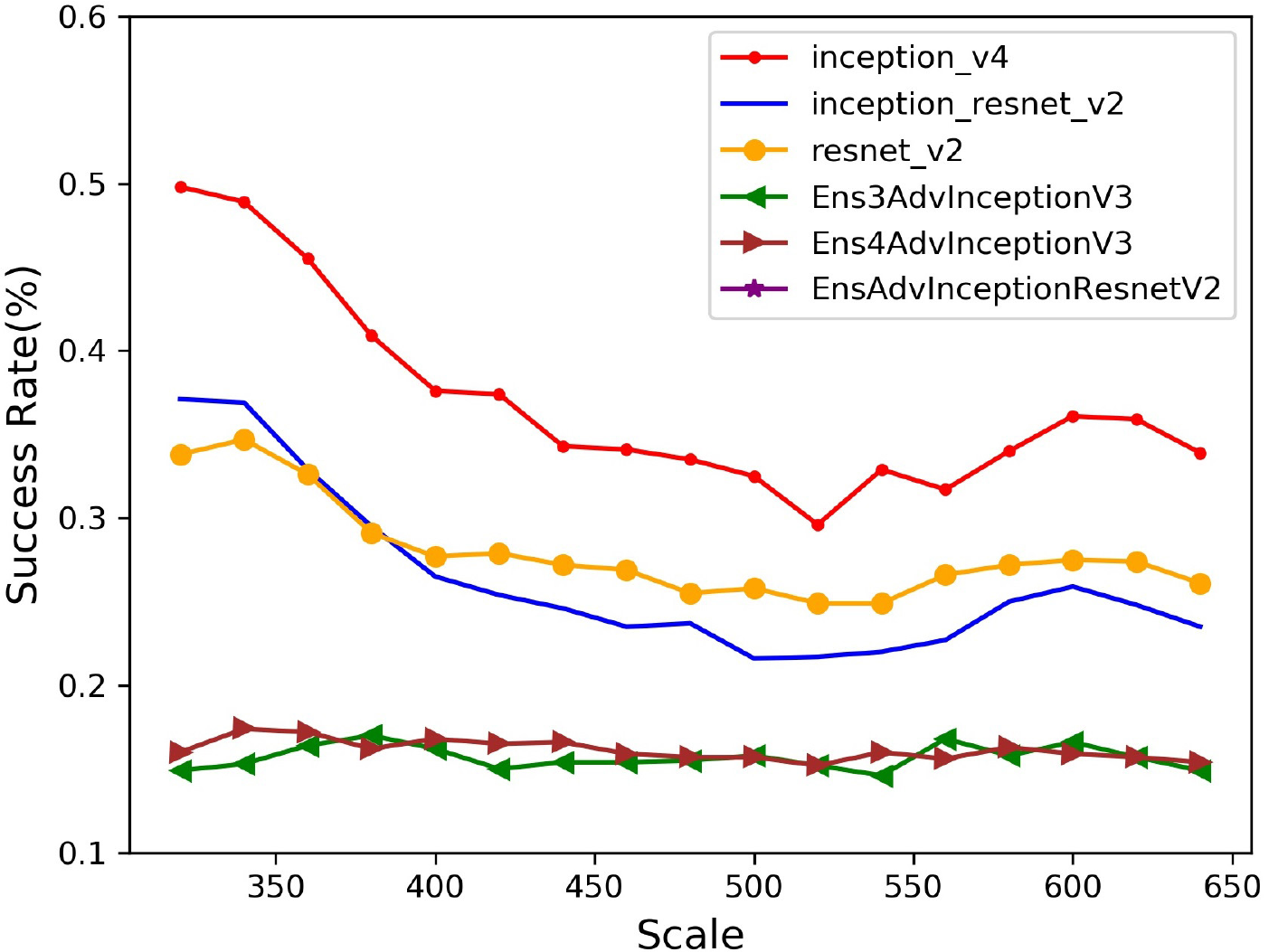}
\end{minipage}
}
\subfigure[RDI-MI-FGSM]{
\begin{minipage}[b]{0.3\linewidth}
\label{fig:3.2}
\includegraphics[width=0.97\textwidth]{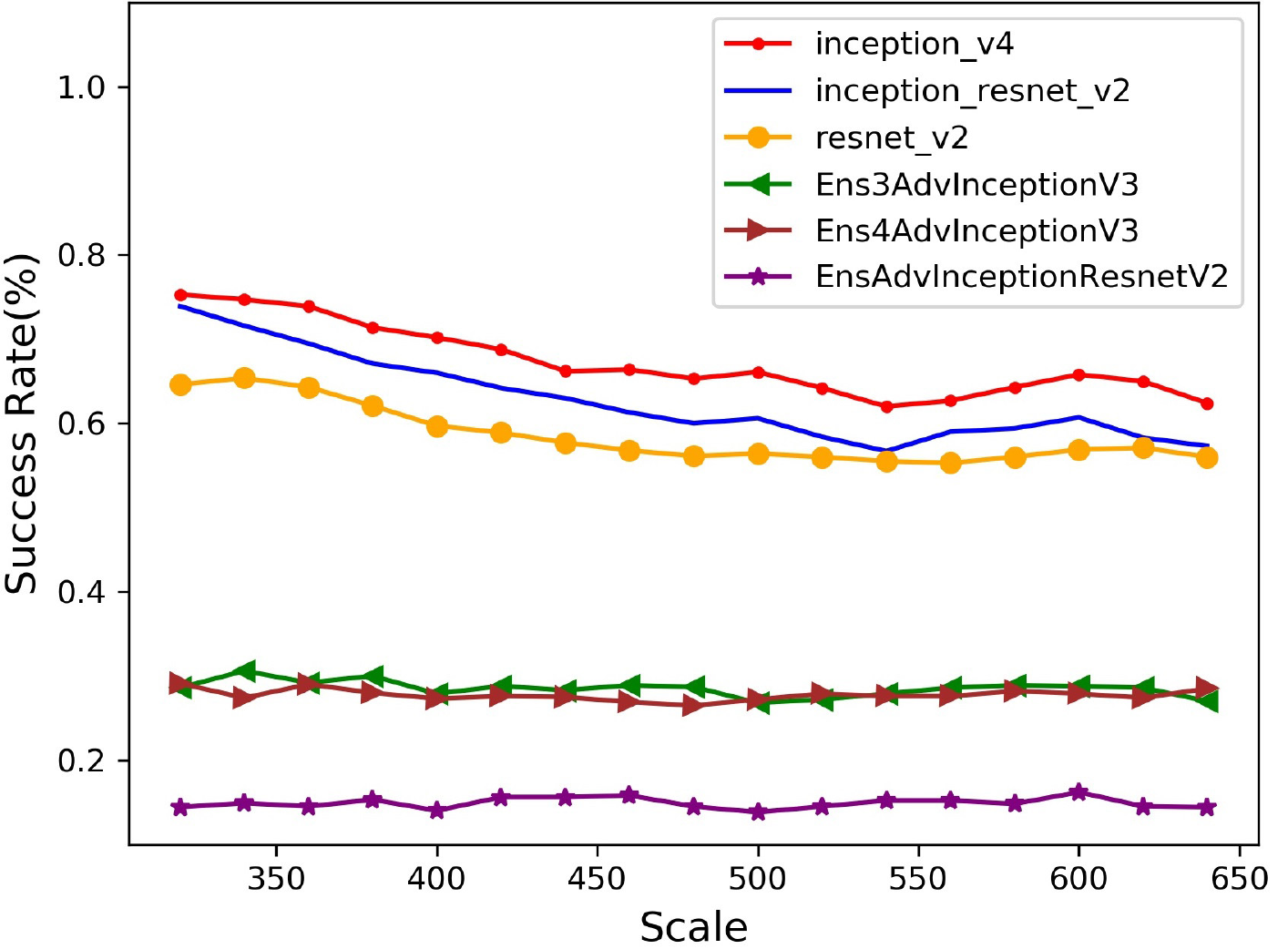}
\end{minipage}
}
\subfigure[TI-RDIM]{
\begin{minipage}[b]{0.3\linewidth}
\label{fig:3.3}
\includegraphics[width=0.97\textwidth]{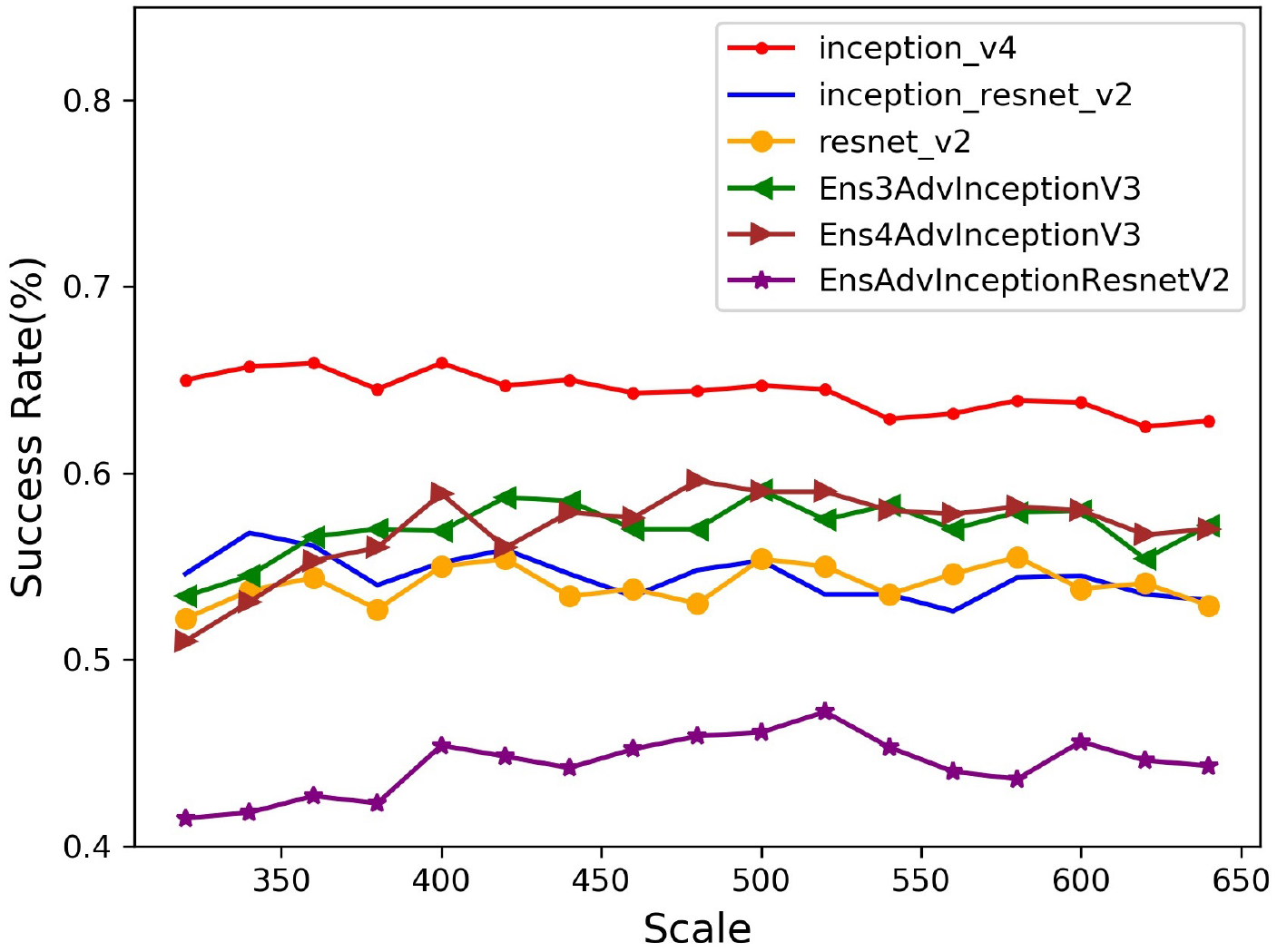}
\end{minipage}
}
 \caption{The success rates (\%) of black-box attacks against six black-box models---Inc-v4, IncRes-v2, Res-v2-152,Inc-v3ens3, Inc-v3ens4 and IncRes-v2ens. The adversarial examples are crafted for Inc-v3 respectively using RDI-FGSM, RDI-MI-FGSM and TI-RDIM with the diversity scale ranging from 320 to 500} \label{fig:4}
\end{figure*}

In this subsection, we attack the Inc-v3 model by RDI-FGSM, RDI-MI-FGSM, and TI-RDIM with different diversity scales and show the success rates against six black-box models in Fig.~\ref{fig:4}. It can be seen that the success rates of RDI-FGSM and RDI-MI-FGSM decrease as diversity scale increasing, while success rates of TI-RDIM continue increasing at first and slightly dropping after the diversity scale exceeds 520.

Based on Fig.~\ref{fig:4}, we further explore the internal relationship between RDIM and TIM. We find that images with vertical and horizontal stripes are more likely to fail when attacking DNNs even if they are perturbed by the translation-invariant method. We present two sets of perturbed images in Fig.~\ref{fig:2}. Additionally, we show gradients of diverse inputs (here we visualize the gradients as images) which have many vertical and horizontal stripes in Fig.~\ref{fig:3}. These three figures indicate that DIM can reduce the effect of stripes on TIM, and thus make adversarial examples generated by the combination of these two methods more transferable. However, without noticing that a certain number of stripes benefit TIM, DIM sets up the transformation probability ${p}$ and limits the maximum diversity scale to 330 to avoid success rates dropping. Hence, we propose a \textbf{resized-diverse-inputs method} (RDIM) by removing the transformation probability ${p}$, setting a much larger diversity size and resizing the diverse inputs to their original size at each iteration. These three groups of interesting results of Fig.~\ref{fig:2}, Fig.~\ref{fig:3} and Fig.~\ref{fig:4} show that RDIM can reduce the effect of stripes on TIM, and thus make adversarial examples generated by the combination of RDIM and TIM more transferable. The experimental results validate the first three analyses of Sec.~\ref{3.2}.

\subsection{Single-Model Attacks}
\label{s4.3}

In this subsection, we categorize the experiments of single-model attacks into two types---single-scale attacks and multi-scale attacks based on time efficiency, e.g., all methods of single-scale attacks have similar runtime in generating adversarial examples. We compare the black-box success rates of the resized-diverse-inputs based methods with single-scale attacks and multi-scale attacks, respectively. In single-scale attacks, we generate adversarial examples for four normally trained models respectively using TI-DIM, TI-RDIM, NI-TI-DIM and NI-TI-RDIM. We then use six defense models to defense the crafted adversarial examples. We present the success rates in Table~\ref{tab:1} for the comparison of TI-DIM and TI-RDIM, and Table~\ref{tab:2} for the comparison of NI-TI-DIM and NI-TI-RDIM.

\begin{table*}[!t]
	\centering
    \renewcommand\arraystretch{.95}
	\caption{\label{tab:1}The success rates (\%) of black-box attacks against six defense models under single-model setting. The adversarial examples are generated for Inc-v3, Inc-v4, IncRes-v2, Res-v2-152 respectively using TI-DIM and TI-RDIM}
	\begin{tabular}{c|c|c|c|c|c|c|c}
		\hline
		 & Attack & Inc-v3ens3 & Inc-v3ens4 & IncRes-v2ens & HGD & R\&P & NIPS-r3\\
		\hline
        \hline
		Inc-v3 & \begin{tabular}[c]{@{}c@{}}TI-DIM\\ TI-RDIM\end{tabular} & \begin{tabular}[c]{@{}c@{}}46.6\\ \textbf{59.1}\end{tabular} & \begin{tabular}[c]{@{}c@{}}47.6\\ \textbf{59.0}\end{tabular} & \begin{tabular}[c]{@{}c@{}}38.1\\ \textbf{46.1}\end{tabular}
		& \begin{tabular}[c]{@{}c@{}}38.1\\ \textbf{48.3}\end{tabular} & \begin{tabular}[c]{@{}c@{}}37.4\\ \textbf{47.5}\end{tabular} & \begin{tabular}[c]{@{}c@{}}42.8\\ \textbf{52.1}\end{tabular}\\
		\hline
		Inc-v4 & \begin{tabular}[c]{@{}c@{}}TI-DIM\\ TI-RDIM\end{tabular} & \begin{tabular}[c]{@{}c@{}}48.2\\ \textbf{61.7}\end{tabular} & \begin{tabular}[c]{@{}c@{}}48.3\\ \textbf{62.0}\end{tabular} & \begin{tabular}[c]{@{}c@{}}39.3\\ \textbf{50.8}\end{tabular}
		& \begin{tabular}[c]{@{}c@{}}41.2\\ \textbf{53.2}\end{tabular} & \begin{tabular}[c]{@{}c@{}}40.7\\ \textbf{51.5}\end{tabular} & \begin{tabular}[c]{@{}c@{}}42.5\\ \textbf{55.7}\end{tabular}\\
		\hline
		IncRes-v2 & \begin{tabular}[c]{@{}c@{}}TI-DIM\\ TI-RDIM\end{tabular} & \begin{tabular}[c]{@{}c@{}}61.3\\ \textbf{69.5}\end{tabular} & \begin{tabular}[c]{@{}c@{}}60.8\\ \textbf{69.0}\end{tabular} & \begin{tabular}[c]{@{}c@{}}59.3\\ \textbf{67.1}\end{tabular}
		& \begin{tabular}[c]{@{}c@{}}59.7\\ \textbf{66.8}\end{tabular} & \begin{tabular}[c]{@{}c@{}}60.9\\ \textbf{67.7}\end{tabular} & \begin{tabular}[c]{@{}c@{}}62.1\\ \textbf{69.7}\end{tabular}\\
		\hline
		Res-v2-152 & \begin{tabular}[c]{@{}c@{}}TI-DIM\\ TI-RDIM\end{tabular} & \begin{tabular}[c]{@{}c@{}}56.2\\ \textbf{61.5}\end{tabular} & \begin{tabular}[c]{@{}c@{}}54.9\\ \textbf{64.1}\end{tabular} & \begin{tabular}[c]{@{}c@{}}50.1\\ \textbf{53.8}\end{tabular}
		& \begin{tabular}[c]{@{}c@{}}52.6\\ \textbf{53.4}\end{tabular} & \begin{tabular}[c]{@{}c@{}}51.1\\ \textbf{52.7}\end{tabular} & \begin{tabular}[c]{@{}c@{}}53.1\\ \textbf{59.0}\end{tabular}\\
		\hline
	\end{tabular}
\end{table*}

\begin{table*}[!t]
	\centering
    \renewcommand\arraystretch{.95}
	\caption{\label{tab:2}The success rates (\%) of black-box attacks against six defense models under single-model setting. The adversarial examples are generated for Inc-v3, Inc-v4, IncRes-v2, Res-v2-152 respectively using NI-TI-DIM and NI-TI-RDIM}
	\begin{tabular}{c|c|c|c|c|c|c|c}
		\hline
		 & Attack & Inc-v3ens3 & Inc-v3ens4 & IncRes-v2ens & HGD & R\&P & NIPS-r3\\
		\hline
        \hline
		Inc-v3 & \begin{tabular}[c]{@{}c@{}}NI-TI-DIM\\ NI-TI-RDIM\end{tabular} & \begin{tabular}[c]{@{}c@{}}50.0\\ \textbf{53.4}\end{tabular} & \begin{tabular}[c]{@{}c@{}}48.7\\ \textbf{52.6}\end{tabular} & \begin{tabular}[c]{@{}c@{}}36.7\\ \textbf{39.8}\end{tabular}
		& \begin{tabular}[c]{@{}c@{}}37.5\\ \textbf{42.3}\end{tabular} & \begin{tabular}[c]{@{}c@{}}36.5\\ \textbf{41.0}\end{tabular} & \begin{tabular}[c]{@{}c@{}}42.6\\ \textbf{46.3}\end{tabular}\\
		\hline
		Inc-v4 & \begin{tabular}[c]{@{}c@{}}NI-TI-DIM\\ NI-TI-RDIM\end{tabular} & \begin{tabular}[c]{@{}c@{}}52.5\\ \textbf{57.9}\end{tabular} & \begin{tabular}[c]{@{}c@{}}52.7\\ \textbf{56.5}\end{tabular} & \begin{tabular}[c]{@{}c@{}}40.1\\ \textbf{45.3}\end{tabular}
		& \begin{tabular}[c]{@{}c@{}}43.2\\ \textbf{48.9}\end{tabular} & \begin{tabular}[c]{@{}c@{}}40.7\\ \textbf{47.6}\end{tabular} & \begin{tabular}[c]{@{}c@{}}42.5\\ \textbf{50.7}\end{tabular}\\
		\hline
		IncRes-v2 & \begin{tabular}[c]{@{}c@{}}NI-TI-DIM\\ NI-TI-RDIM\end{tabular} & \begin{tabular}[c]{@{}c@{}}61.1\\ \textbf{66.1}\end{tabular} & \begin{tabular}[c]{@{}c@{}}60.2\\ \textbf{65.5}\end{tabular} & \begin{tabular}[c]{@{}c@{}}60.3\\ \textbf{62.8}\end{tabular}
		& \begin{tabular}[c]{@{}c@{}}60.7\\ \textbf{65.8}\end{tabular} & \begin{tabular}[c]{@{}c@{}}61.2\\ \textbf{64.3}\end{tabular} & \begin{tabular}[c]{@{}c@{}}62.7\\ \textbf{66.0}\end{tabular}\\
		\hline
		Res-v2-152 & \begin{tabular}[c]{@{}c@{}}NI-TI-DIM\\ NI-TI-RDIM\end{tabular} & \begin{tabular}[c]{@{}c@{}}56.1\\ \textbf{60.1}\end{tabular} & \begin{tabular}[c]{@{}c@{}}55.9\\ \textbf{60.4}\end{tabular} & \begin{tabular}[c]{@{}c@{}}51.2\\ \textbf{59.9}\end{tabular}
		& \begin{tabular}[c]{@{}c@{}}50.1\\ \textbf{52.4}\end{tabular} & \begin{tabular}[c]{@{}c@{}}49.1\\ \textbf{51.4}\end{tabular} & \begin{tabular}[c]{@{}c@{}}53.7\\ \textbf{57.6}\end{tabular}\\
		\hline
	\end{tabular}
\end{table*}

It can be observed from the tables that our method RDIM can further boost the success rates against these six defense models by a large margin when integrated into the state-of-the-art attacks. In general, the resized-diverse-inputs based methods outperform the baseline methods by $2\% \sim 14\%$. It demonstrates that our method RDIM is better than DIM, and can serve as a powerful method to improve the transferability of adversarial examples.

In multi-scale attacks, we also generate adversarial examples for four normally trained models respectively using SI-TIM, DE-TIM, SI-NI-TIM and DE-NI-TIM. We then evaluate the crafted adversarial examples by attacking six defense models. We present the success rates in Table~\ref{tab:3} for the comparison of SI-TIM and DE-TIM. Table~\ref{tab:4} presents the comparison of SI-NI-TIM and DE-NI-TIM.

\begin{table*}[!t]
	\centering
    \renewcommand\arraystretch{.95}
	\caption{\label{tab:3}The success rates (\%) of black-box attacks against six defense models under single-model setting. The adversarial examples are generated for Inc-v3, Inc-v4, IncRes-v2, Res-v2-152 respectively using SI-TIM and DE-TIM}
	\begin{tabular}{c|c|c|c|c|c|c|c}
		\hline
		 & Attack & Inc-v3ens3 & Inc-v3ens4 & IncRes-v2ens & HGD & R\&P & NIPS-r3\\
		\hline
        \hline
		Inc-v3 & \begin{tabular}[c]{@{}c@{}}SI-TIM\\ DE-TIM\end{tabular} & \begin{tabular}[c]{@{}c@{}}48.4\\ \textbf{70.1}\end{tabular} & \begin{tabular}[c]{@{}c@{}}51.2\\ \textbf{70.3}\end{tabular} & \begin{tabular}[c]{@{}c@{}}37.5\\ \textbf{58.0}\end{tabular}
		& \begin{tabular}[c]{@{}c@{}}36.3\\ \textbf{61.2}\end{tabular} & \begin{tabular}[c]{@{}c@{}}34.6\\ \textbf{59.3}\end{tabular} & \begin{tabular}[c]{@{}c@{}}40.0\\ \textbf{64.2}\end{tabular}\\
		\hline
		Inc-v4 & \begin{tabular}[c]{@{}c@{}}SI-TIM\\ DE-TIM\end{tabular} & \begin{tabular}[c]{@{}c@{}}51.2\\ \textbf{71.1}\end{tabular} & \begin{tabular}[c]{@{}c@{}}50.9\\ \textbf{69.2}\end{tabular} & \begin{tabular}[c]{@{}c@{}}42.9\\ \textbf{59.6}\end{tabular}
		& \begin{tabular}[c]{@{}c@{}}41.9\\ \textbf{64.2}\end{tabular} & \begin{tabular}[c]{@{}c@{}}39.5\\ \textbf{63.4}\end{tabular} & \begin{tabular}[c]{@{}c@{}}42.5\\ \textbf{65.1}\end{tabular}\\
		\hline
		IncRes-v2 & \begin{tabular}[c]{@{}c@{}}SI-TIM\\ DE-TIM\end{tabular} & \begin{tabular}[c]{@{}c@{}}68.8\\ \textbf{79.8}\end{tabular} & \begin{tabular}[c]{@{}c@{}}66.1\\ \textbf{79.5}\end{tabular} & \begin{tabular}[c]{@{}c@{}}65.4\\ \textbf{78.2}\end{tabular}
		& \begin{tabular}[c]{@{}c@{}}60.6\\ \textbf{80.0}\end{tabular} & \begin{tabular}[c]{@{}c@{}}59.4\\ \textbf{79.3}\end{tabular} & \begin{tabular}[c]{@{}c@{}}62.7\\ \textbf{80.1}\end{tabular}\\
		\hline
		Res-v2-152 & \begin{tabular}[c]{@{}c@{}}SI-TIM\\ DE-TIM\end{tabular} & \begin{tabular}[c]{@{}c@{}}54.7\\ \textbf{77.5}\end{tabular} & \begin{tabular}[c]{@{}c@{}}55.3\\ \textbf{75.8}\end{tabular} & \begin{tabular}[c]{@{}c@{}}48.0\\ \textbf{69.4}\end{tabular}
		& \begin{tabular}[c]{@{}c@{}}45.2\\ \textbf{73.9}\end{tabular} & \begin{tabular}[c]{@{}c@{}}43.4\\ \textbf{71.8}\end{tabular} & \begin{tabular}[c]{@{}c@{}}48.4\\ \textbf{75.0}\end{tabular}\\
		\hline
	\end{tabular}
\end{table*}

\begin{table*}[!t]
	\centering
    \renewcommand\arraystretch{.95}
	\caption{\label{tab:4}The success rates (\%) of black-box attacks against six defense models under single-model setting. The adversarial examples are generated for Inc-v3, Inc-v4, IncRes-v2, Res-v2-152 respectively using SI-NI-TIM and DE-NI-TIM}
	\begin{tabular}{c|c|c|c|c|c|c|c}
		\hline
		 & Attack & Inc-v3ens3 & Inc-v3ens4 & IncRes-v2ens & HGD & R\&P & NIPS-r3\\
		\hline
        \hline
		Inc-v3 & \begin{tabular}[c]{@{}c@{}}SI-NI-TIM\\ DE-NI-TIM\end{tabular} & \begin{tabular}[c]{@{}c@{}}52.1\\ \textbf{66.4}\end{tabular} & \begin{tabular}[c]{@{}c@{}}52.8\\ \textbf{66.8}\end{tabular} & \begin{tabular}[c]{@{}c@{}}40.7\\ \textbf{52.7}\end{tabular}
		& \begin{tabular}[c]{@{}c@{}}39.5\\ \textbf{56.2}\end{tabular} & \begin{tabular}[c]{@{}c@{}}37.3\\ \textbf{55.4}\end{tabular} & \begin{tabular}[c]{@{}c@{}}44.4\\ \textbf{59.2}\end{tabular}\\
		\hline
		Inc-v4 & \begin{tabular}[c]{@{}c@{}}SI-NI-TIM\\ DE-NI-TIM\end{tabular} & \begin{tabular}[c]{@{}c@{}}55.6\\ \textbf{67.3}\end{tabular} & \begin{tabular}[c]{@{}c@{}}54.1\\ \textbf{65.2}\end{tabular} & \begin{tabular}[c]{@{}c@{}}44.7\\ \textbf{56.4}\end{tabular}
		& \begin{tabular}[c]{@{}c@{}}43.1\\ \textbf{60.9}\end{tabular} & \begin{tabular}[c]{@{}c@{}}41.4\\ \textbf{59.2}\end{tabular} & \begin{tabular}[c]{@{}c@{}}46.3\\ \textbf{62.7}\end{tabular}\\
		\hline
		IncRes-v2 & \begin{tabular}[c]{@{}c@{}}SI-NI-TIM\\ DE-NI-TIM\end{tabular} & \begin{tabular}[c]{@{}c@{}}68.6\\ \textbf{77.5}\end{tabular} & \begin{tabular}[c]{@{}c@{}}66.5\\ \textbf{75.5}\end{tabular} & \begin{tabular}[c]{@{}c@{}}64.1\\ \textbf{74.2}\end{tabular}
		& \begin{tabular}[c]{@{}c@{}}57.9\\ \textbf{75.1}\end{tabular} & \begin{tabular}[c]{@{}c@{}}58.4\\ \textbf{76.2}\end{tabular} & \begin{tabular}[c]{@{}c@{}}61.9\\ \textbf{77.9}\end{tabular}\\
		\hline
		Res-v2-152 & \begin{tabular}[c]{@{}c@{}}SI-NI-TIM\\ DE-NI-TIM\end{tabular} & \begin{tabular}[c]{@{}c@{}}57.6\\ \textbf{74.5}\end{tabular} & \begin{tabular}[c]{@{}c@{}}55.8\\ \textbf{74.8}\end{tabular} & \begin{tabular}[c]{@{}c@{}}48.7\\ \textbf{67.5}\end{tabular}
		& \begin{tabular}[c]{@{}c@{}}47.9\\ \textbf{69.3}\end{tabular} & \begin{tabular}[c]{@{}c@{}}46.2\\ \textbf{68.6}\end{tabular} & \begin{tabular}[c]{@{}c@{}}53.3\\ \textbf{73.0}\end{tabular}\\
		\hline
	\end{tabular}
\end{table*}

We can observe from the tables that our method DEM can further improve the success rates against these six defense models by a large margin when integrated into the state-of-the-art attacks. In general, methods combined with DEM outperform the baseline methods by $11\% \sim 24\%$. In particular, when using DE-TIM, the combination of our method and TIM, to attack IncRes-v2 model, the adversarial examples achieve no less than 78\% success rates against all six defense models. In Table~\ref{tab:1}, Table~\ref{tab:2}, Table~\ref{tab:3} and Table~\ref{tab:4}, it should be noted that the adversarial examples crafted for a non-defense model can fool six defense models with no less than 78\% success rates. The results not only validate the effectiveness of RDIM and DEM, but also indicate that the current defenses fail to meet the demand of practical security.

\subsection{Ensemble-based Attacks}
\label{4.4}

In the subsection, we further show the performance of adversarial examples crafted for an ensemble of models. Similar to Sec.~\ref{s4.3}, we categorize the experiments of Ensemble-based attacks into single-scale attacks and multi-scale attacks. We generate adversarial examples for the ensemble of Inc-v3, Inc-v4, IncRes-v2 and Res-v2-152 with equal ensemble weights.

In single-scale attacks, we generate adversarial examples respectively using TI-DIM, TI-RDIM, NI-TI-DIM and RF-TI-RDIM, and evaluate the effectiveness of crafted adversarial examples by attacking six defenses. Table~\ref{tab:5} shows the results of black-box attacks against six defenses. The results indicate that the proposed method RDIM can also boost the success rates over the baselines attacks in ensemble-based attacks.

\begin{table*}[!t]
\renewcommand\arraystretch{.95}
\caption{\label{tab:5}The success rates (\%) of black-box attacks against six defense models under multi-model setting. The adversarial examples are generated for the ensemble of Inc-v3, Inc-v4, IncRes-v2, Res-v2-152 using TI-DIM, TI-RDIM, NI-TI-DIM and RF-TI-RDIM}
\begin{center}
\begin{tabular}{c|c|c|c|c|c|c}
\hline
Attack & Inc-v3ens3 & Inc-v3ens4 & IncRes-v2ens & HGD & R\&P & NIPS-r3\\
\hline\hline
TI-DIM & 83.8& 83.1& 78.5&83.0 &81.7 &83.7\\
TI-RDIM &85.0&84.9&79.1&82.1&81.2&83.9\\
NI-TI-DIM &86.4&84.9&79.4&82.3&81.0&84.2\\
RF-TI-RDIM &\textbf{91.3}&\textbf{90.1}&\textbf{82.0}&\textbf{87.9}&\textbf{86.1}&\textbf{90.7}\\
\hline
\end{tabular}
\end{center}
\end{table*}

\begin{table*}[!t]
\renewcommand\arraystretch{.95}
\caption{\label{tab:6}The success rates (\%) of black-box attacks against six defense models under multi-model setting. The adversarial examples are generated for the ensemble of Inc-v3, Inc-v4, IncRes-v2, Res-v2-152 using SI-NI-TIM, SI-NI-TI-DIM, DE-NI-TIM, DE-TIM and RF-DE-TIM}
\begin{center}
\begin{tabular}{c|c|c|c|c|c|c}
\hline
Attack & Inc-v3ens3 & Inc-v3ens4 & IncRes-v2ens & HGD & R\&P & NIPS-r3\\
\hline\hline
SI-NI-TIM & 79.5& 79.1& 70.3&73.4 &71.5 &77.2\\
SI-NI-TI-DIM &87.2&85.6&77.7&82.3&81.6&84.5\\
DE-NI-TIM &81.5&79.6&69.8&76.1&74.8&78.6\\
DE-TIM &91.2&90.7&88.2&90.5&90.1&91.1\\
RF-DE-TIM &\textbf{94.7}&\textbf{94.5}&\textbf{89.1}&\textbf{93.2}&\textbf{92.7}&\textbf{93.9}\\
\hline
\end{tabular}
\end{center}
\end{table*}

In multi-scale attacks, we further present adversarial examples respectively using SI-NI-TIM, SI-NI-TI-DIM, DE-NI-TIM, DE-TIM and RF-DE-TIM, and then employ six defense models to defense the generated adversarial examples. Table~\ref{tab:6} shows that our method DEM and region fitting can be easily integrated into state-of-the-art attack methods and improve the transferability of adversarial examples. The experimental results prove the fourth analysis in Sec.~\ref{3.2}. In particular, our best attack RF-DE-TIM fools six defense models with a 93\% success rate on average. Such high success rates mean that there is an urgent need to develop more defensive methods to resist adversarial examples.

\section{Conclusion}
\label{conclusion}

In this paper, we introduce a three stage pipeline: resized-diverse-inputs (RDIM), diversity-ensemble (DEM) and region fitting, that work together to generate transferable adversarial examples. We first explore the internal relationship between DIM and TIM, and propose RDIM that is more suitable to characterize this relationship. Combined with TIM, RDIM can balance the contradiction between loss of gradient information and stripes demand. Then we propose DEM, the multi-scale version of RDIM, to generate multi-scale gradients with different numbers of vertical and horizontal stripes for TIM. After the first two steps we transform value fitting into region fitting across iterations. RDIM and region fitting do not require extra running time and these three steps can be well integrated into other attacks. Our best attack RF-DE-TIM fools six black-box defenses with a 93\% attack success rate on average, which is higher than the state-of-the-art multi-model attacks. We hope that our findings about attack methods will shed light into potential future directions for adversarial attacks.

%
%
\bibliographystyle{splncs04}
\bibliography{eccv2020submissionCR}
\end{document}